\pdfoutput=1
\documentclass[11pt]{article}

\usepackage[]{acl}

\usepackage{times}
\usepackage{latexsym}

\usepackage{xcolor}
\usepackage{lipsum}

\usepackage[T1]{fontenc}

\usepackage[utf8]{inputenc}

\usepackage{microtype}

\usepackage{graphicx}
\usepackage{adjustbox}
\usepackage{amsmath}
\usepackage{amsfonts}
\usepackage{algorithm}
\usepackage{bm}
\usepackage{soul}
\usepackage[noend]{algpseudocode}
\usepackage[colorinlistoftodos]{todonotes}

\algrenewcommand\algorithmicindent{1.0em}
\algnewcommand{\LineComment}[1]{\State \(\triangleright\) #1}
\usepackage{amsthm}

\def\parsum#1{\bgroup \textcolor{blue}{Paragraph summary: #1}\egroup}

\def\bluenote#1{\bgroup \textcolor{blue}{#1} \egroup}
\def\cyannote#1{\bgroup \textcolor{cyan}{#1} \egroup}

\newcommand{\name}{\texttt{DP-NMT}}

\title{DP-NMT: Scalable Differentially-Private Machine Translation}

\author{Timour Igamberdiev\textsuperscript{$1$} \,
	Doan Nam Long Vu\textsuperscript{$1$} \,
	Felix K\"unnecke\textsuperscript{$1$} \\
	\textbf{Zhuo Yu}\textsuperscript{$1$} \,
	\textbf{Jannik Holmer}\textsuperscript{$1$} \,
	\textbf{Ivan Habernal}\textsuperscript{$2$} \\
	Trustworthy Human Language Technologies \\
	\textsuperscript{$1$} Department of Computer Science, Technical University of Darmstadt \\
	\textsuperscript{$2$} Department of Computer Science, Paderborn University \\
	{\texttt{timour.igamberdiev@tu-darmstadt.de}} \\
	\url{www.trusthlt.org}
}

\begin{document}
    \onecolumn
    \noindent \textbf{DP-NMT: Scalable Differentially-Private Machine Translation}

    \medskip
    \noindent Timour Igamberdiev, Doan Nam Long Vu, Felix K\"unnecke, Zhuo Yu, Jannik Holmer and Ivan Habernal

    \bigskip
    This is a \textbf{camera-ready version} of the article accepted for publication at the \emph{18th Conference of the European Chapter of the Association for Computational Linguistics (EACL 2024)}. The final official version is published on the ACL Anthology website: \url{https://aclanthology.org/}

    \medskip
    Please cite this pre-print version as follows.
    \medskip

\begin{verbatim}
@InProceedings{Igamberdiev.2024.EACL,
    title = {DP-NMT: Scalable Differentially-Private
             Machine Translation},
    author = {Igamberdiev, Timour and
              Vu, Doan Nam Long and
              Kuennecke, Felix and
              Yu, Zhuo and
              Holmer, Jannik and
              Habernal, Ivan},
    publisher = {Association for Computational Linguistics},
    booktitle = {Proceedings of the 18th Conference
                 of the European Chapter of the 
                 Association for Computational Linguistics:
                 System Demonstrations},
    pages = {94--105},
    year = {2024},
    address = {St. Julians, Malta}
}
\end{verbatim}
    \twocolumn

\maketitle
\begin{abstract}
Neural machine translation (NMT) is a widely popular text generation task, yet there is a considerable research gap in the development of privacy-preserving NMT models, despite significant data privacy concerns for NMT systems.
Differentially private stochastic gradient descent (DP-SGD) is a popular method for training machine learning models with concrete privacy guarantees; however, the implementation specifics of training a model with DP-SGD are not always clarified in existing models, 
with differing software libraries used and code bases not always being public, leading to reproducibility issues.
To tackle this, we introduce \name, an open-source framework for carrying out research on privacy-preserving NMT with DP-SGD, bringing together numerous models, datasets, and evaluation metrics in one systematic software package. 
Our goal is to provide a platform for researchers to advance the development of privacy-preserving NMT systems, keeping the specific details of the DP-SGD algorithm transparent and intuitive to implement. 
We run a set of experiments on datasets from both general and privacy-related domains to demonstrate our framework in use. 
We make our framework publicly available and welcome feedback from the community.\footnote{\url{https://github.com/trusthlt/dp-nmt}}
\end{abstract}

\section{Introduction}

\setstcolor{blue}

Privacy-preserving natural language processing (NLP) has been a recently growing field, in large part due to an increasing amount of concern regarding data privacy. This is especially a concern in the context of modern neural networks memorizing training data that may contain sensitive information \citep{carlini2021extracting}.
While there has been a body of research investigating privacy for text classification tasks \citep{Senge.et.al.2022.EMNLP} and language models \citep{Hoory.et.al.2021.FindingsEMNLP,anil-etal-2022-large}, there has not been as much focus on text generation tasks, in particular neural machine translation (NMT).
However, NMT is particularly worrying from a privacy perspective, due to a variety of machine translation services available online that users send their personal data to.
This includes built-in NMT services to existing websites, e-mail clients, and search engines. 
After data has been sent to these systems, it may be further processed and used in the development of the NMT system \citep{kamocki2016privacy}, which has a significant risk of being memorized if trained in a non-private manner.

One of the most popular methods for tackling this privacy issue is differential privacy (DP), being a formal framework which provides probabilistic guarantees that the contribution of any single data point to some analysis is bounded.
In the case of NLP and machine learning (ML), this means that a data point associated with some individual which is included in the model's training data cannot stand out `too much' in the learning process of the model.

The DP-SGD algorithm \citep{abadi2016deep} is one of the most standard methods to achieve this for ML systems, yet implementations of DP-SGD often lack some technical details on the specifics of the algorithm.
In particular, this includes the privacy amplification method assumed for calculating the privacy budget $\varepsilon$ when composed over all training iterations of the model.
This means that the exact \textit{strength of the privacy protection} that the resulting systems provide is not clear, with the `standard' \textbf{random shuffling} method for iterating over batches providing a weaker privacy guarantee for the training data than \textbf{Poisson sampling}. 
With different implementations using different software libraries, the community currently does not have a consistent platform for conducting experiments for scalable differentially private systems, such as NMT. 

To tackle this problem, we develop a modular framework for conducting research on private NMT in a transparent and reproducible manner.
Our primary goal is to allow for a deeper investigation into the applications of DP for NMT, all while ensuring that important theoretical details of the DP-SGD methodology are properly reflected in the implementation.
Following previous work on DP-SGD \citep{subramani2021enabling,anil-etal-2022-large}, we implement our framework in the JAX library \citep{jax2018github}, which provides powerful tools that help to reduce the significant computational overhead of DP-SGD, allowing for scalability in implementing larger systems and more extended training regimes.

Our primary contributions are as follows.
First, we present \name, a framework developed in JAX for leading research on NMT with DP-SGD. It includes a growing list of available NMT models, different evaluation schemes, as well as numerous datasets available out of the box, including standard datasets used for NMT research and more specific privacy-related domains.
Second, we demonstrate our framework by running experiments on these NMT datasets, providing one of the first investigations into privacy-preserving NMT.
Importantly, we compare the random shuffling and Poisson sampling methods for iterating over training data when using DP-SGD.
We demonstrate that, in addition to the theoretical privacy guarantee, there may indeed be differences in the model performance when utilizing each of the two settings.

\section{DP-SGD and subsampling}
\label{sec:dp-sgd-subsampling}

We describe the main ideas of differential privacy (DP) and DP-SGD in Appendix~\ref{sec:appx-dp-background}.
We refer to \citet{abadi2016deep,Igamberdiev.Habernal.2022.LREC,Habernal.2021.EMNLP,Habernal.2022.ACL,Hu.et.al.2024.EACL} for a more comprehensive explanation.

A key aspect of the DP-SGD algorithm (see Alg.~\ref{alg:dp-sgd} in the Appendix) is \textbf{privacy amplification by subsampling}, in which a stronger privacy guarantee can be obtained for a given dataset $x$ when a subset of this dataset is first randomly sampled \citep{kasiviswanathan2011can,beimel2014bounds}.
If the sampling probability is $q$, then the overall privacy guarantee can be analyzed as being approximately $q\varepsilon$.

A key point here is the nature of this sampling procedure and the resulting privacy guarantee.
The moments accountant of \citet{abadi2016deep}, which is an improvement on the strong composition theorem \citep{dwork2010boosting} for composing multiple DP mechanisms, assumes Poisson sampling.
Under this procedure, \textit{each data point} is included in a mini-batch with probability $q = L/N$, with $L$ being the \textit{lot size} and $N$ the size of the dataset.
An alternative method to Poisson sampling is uniform sampling, in which mini-batches of a fixed size are independently drawn at each training iteration \citep{wang2019subsampled,balle2018privacy}.

In practice, however, many modern implementations of DP-SGD utilize \textbf{random shuffling}, with the dataset split into fixed-size mini-batches.
Several training iterations thus form an epoch, in which each training data point appears exactly once, in contrast to Poisson sampling for which the original notion of `epoch' is not quite suitable, since each data point can appear in any training iteration and there is no ``single passing of the training data through the model''.
In \citet{abadi2016deep}, the term \textit{epoch} is redefined as $\frac{N}{L}$ lots, being essentially an expectation of the number of batches when utilizing $N$ data points for training the model.
While simply shuffling the dataset can indeed result in privacy amplification \citep{erlingsson2019amplification,feldman2022hiding}, the nature of the corresponding privacy guarantee is \textbf{not the same} as the guarantee achieved by Poisson sampling, generally being weaker. We refer to \citet[Section 4.3]{ponomareva2023dp} for further details.

\section{Related work}

\subsection{Applications of DP-SGD to NLP}

The application of DP-SGD to the field of NLP has seen an increasing amount of attention in recent years.
A large part of these studies focus on differentially private pre-training or fine-tuning of language models \citep{Hoory.et.al.2021.FindingsEMNLP,yu2021differentially,basu-etal-2021-privacy,xu-etal-2021-mitigating-data,anil-etal-2022-large,ponomareva-etal-2022-training,shi-etal-2022-selective,wu-etal-2022-adaptive,li2022large,yin-habernal-2022-privacy,mattern2022differentially,hansen2022impact,Senge.et.al.2022.EMNLP}. A primary goal is to reach the best possible privacy/utility trade-off for the trained models, in which the highest performance is achieved with the strictest privacy guarantees.

In the general machine learning setting, the exact sampling method that is used for selecting batches at each training iteration is often omitted, since this is generally not a core detail of the training methodology.
Possibly for this reason, in the case of privately training a model with DP-SGD, the sampling method is also often not mentioned.
However, in contrast to the non-private setting, here \textbf{sampling is actually a core detail of the algorithm}, which has an \textbf{impact on the privacy accounting procedure}.
In the case that experimental descriptions with DP-SGD include mentions of \textit{epochs} without further clarification, this in fact suggests the use of the random shuffling scheme, as opposed to Poisson sampling, as described in Section~\ref{sec:dp-sgd-subsampling}. 
In addition, sometimes the code base is not publicly available, in which case it is not possible to validate the sampling scheme used. 

Finally, standard implementations of DP-SGD in the Opacus \citep{opacus} and TensorFlow Privacy \citep{abadi2016tensorflow} libraries often include descriptions of DP-SGD implementations with randomly shuffled fixed-size batches.
For instance, while Opacus currently has a \texttt{DPDataLoader} class which by default uses their \texttt{UniformWithReplacementSampler} class for facilitating the use of Poisson sampling, some of the tutorials currently offered appear to also use static batches instead.\footnote{\url{https://opacus.ai/tutorials/building_image_classifier}.}
A similar situation is true for TensorFlow Privacy.\footnote{\url{https://www.tensorflow.org/responsible_ai/privacy/tutorials/classification_privacy}.}
While these libraries support per-example gradients as well, several core features of JAX make it the fastest and most scalable option for implementing DP-SGD \citep{subramani2021enabling}, described in more detail below in Section~\ref{sec:methods}.

We therefore stress the importance of clarifying implementation details that may not be as vital in the general machine learning setting, but are very relevant in the private setting. As described by \citet{ponomareva2023dp}, it is an open theoretical question as to how random shuffling and Poisson sampling differ with respect to privacy amplification gains, with known privacy guarantees being weaker for the former.

\subsection{Private neural machine translation}
The task of private neural machine translation remains largely unexplored, with currently no studies we could find that incorporate DP-SGD to an NMT system.
\citet{wang2021modeling} investigate NMT in a federated learning setup \citep{mcmahan2017communication}, with differential privacy included in the aggregation of parameters from each local model, adding Laplace noise to these parameters. 
Several other studies explore NMT with federated learning, but do not incorporate differential privacy in the methodology \citep{roosta2021communication,passban2022training,du2022federated}.
\citet{hisamoto2020membership}, applied a membership inference attack \citep{shokri2017membership} on a 6-layer Transformer \citep{vaswani2017attention} model in the scenario of NMT as a service, with the goal of clients being able to verify whether their data was used to train an NMT model.
Finally, \citet{kamocki2016privacy} address the general topic of privacy issues for machine translation as a service.
The authors examine how these MT services fit European data protection laws, noting the legal nature of various types of data processing that can occur by both the provider of such a service, as well as by the users themselves.

\section{Description of software}
\label{sec:methods}

\begin{figure*}[!ht]
    \centering
    \includegraphics[width=\linewidth]{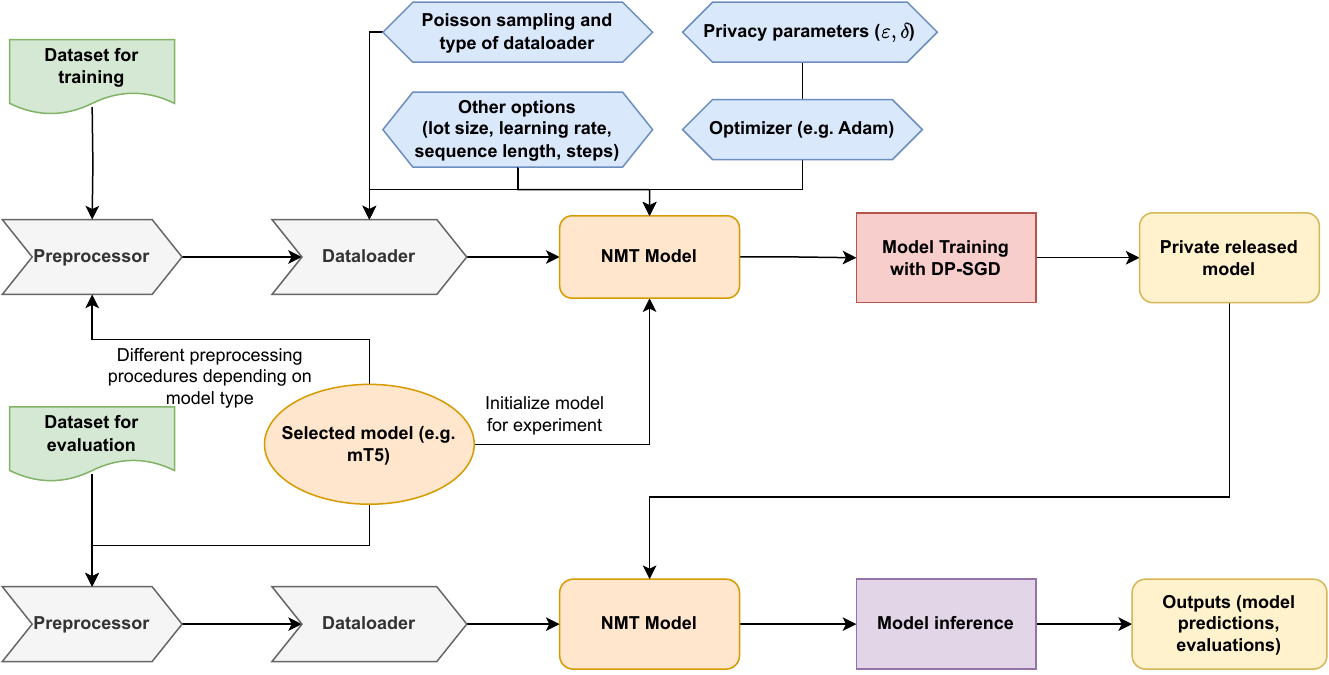}
    \caption{Framework Pipeline. Similar components are represented with different colors. Green: Dataset selection. Blue: Experimental configurations (including privacy settings). Grey: Dataset preparation. Orange: Model-specific elements. Red: Model training. Purple: Model inference. Yellow: Output of experiments.}
    \label{fig:framework-pipeline}
\end{figure*}

The aim of our system is to offer a reliable and scalable approach to achieve differentially private machine translation. Figure \ref{fig:framework-pipeline} illustrates the central structure of our system.
The user can upload a translation dataset that is either accessible on the HuggingFace Datasets Hub\footnote{\url{https://huggingface.co/datasets}} or is provided by us out of the box, and integrate it seamlessly for both training and efficient privacy accounting, utilizing HuggingFace’s Datasets library \cite{lhoest-etal-2021-datasets}. 
\paragraph{Accelerated DP-SGD with JAX and Flax}
Our goal is to accelerate DP-SGD training through the use of a Transformer model implemented with JAX and Flax \cite{jax2018github, flax2020github}. The speed of training DP-SGD in the framework can be considerably enhanced through vectorization, just-in-time (JIT) compilation, and static graph optimization \cite{subramani2021enabling}. JIT compilation and automatic differentiation are defined and established on the XLA compiler. JAX's main transformation methods of interest for fast DP-SGD are \texttt{grad}, \texttt{vmap}, and \texttt{pmap}, offering the ability to mix these operations as needed \cite{yin-habernal-2022-privacy}. In the DP-SGD scenario, combining \texttt{grad} and \texttt{vmap} facilitates efficient computation of per-example gradients by vectorizing the gradient calculation along the batch dimension \cite{anil-etal-2022-large}. Additionally, our training step is decorated by \texttt{pmap} to leverage the XLA compiler on multiple GPUs, significantly accelerating training speed. The framework offers to conduct experiments with multiple encoder-decoder models and integrate new seq2seq models, in addition to existing ones, such as mBART \cite{liu2020multilingual}, T5 \cite{colin-2020-t5}, and mT5 \cite{xue-etal-2021-mt5}. When selecting a model, the corresponding preprocessor will prepare the dataset accordingly. This allows the software to be flexible and modular, enabling researchers to exchange models and datasets to perform a range of private NMT experiments.

\paragraph{Model training and inference}
The experimental workflow of our framework works in two phases, namely model training and model inference. For both phases, the process begins with a data loader that can be either a framework-provided dataset or a user-specified dataset.
Subsequently, the loaded dataset is prepared based on user-defined parameters, including standard options (e.g.\ sequence length), as well as parameters relating to DP-SGD (e.g.\ data loader type, sampling method, and batch size). 
After selecting the model, the user separates it into different procedures according to the model type. Subsequently, the model is initiated, optionally from a checkpoint that has already been trained. Then, the primary experiment is carried out based on the specified mode, which includes (1) fine-tuning on an existing dataset, (2) using an existing fine-tuned checkpoint to continue fine-tuning on the dataset, or (3) inference without teacher forcing.
\paragraph{Integrating \texttt{DPDataloader} from Opacus}
One notable improvement in our software is the incorporation of the \texttt{DPDataloader} from Opacus \cite{opacus} for out-of-the-box Poisson sampling.
This is different from the existing approaches in JAX used by \citet{yin-habernal-2022-privacy, subramani2021enabling, ponomareva-etal-2022-training}, who employ iteration over a randomly shuffled dataset, which theoretically provides weaker DP bounds. Evaluation metrics such as BLEU \cite{papineni2002bleu} and BERTScore \cite{zhang2019bertscore} are available for each mode.
We incorporate the differential privacy component during the training phase of the systems.
\paragraph{Engineering challenges for LLMs}
Throughout development, we encountered multiple engineering challenges. Initially, our academic budget limitations made it difficult to train a larger model due to the significant memory consumption during per-example gradient calculations. Consequently, we anticipated a relatively small physical batch size on each GPU. We attempted to freeze parts of the model for faster training and improved memory efficiency, as \citet{Senge.et.al.2022.EMNLP} noted. However, in Flax, the freezing mechanism only occurs during the optimization step and does not affect per-example gradient computation. Therefore, it does not solve the issue of limited physical batch sizes. Multiple reports suggest that increasing the lot size leads to better DP-SGD performance due to an improved gradient signal-to-noise ratio and an increased likelihood of non-duplicated example sampling across the entire dataset \cite{Hoory.et.al.2021.FindingsEMNLP, yin-habernal-2022-privacy, anil-etal-2022-large}.
However, compared to previous work on large models that mostly relied on dataset iteration \citep{yin-habernal-2022-privacy, ponomareva-etal-2022-training}, implementing the original DP-SGD with large lots using Poisson sampling, a large language model (LLM) with millions of parameters, and on multiple GPUs presents a challenge that makes comparison difficult.
To address this issue, we first conduct a sampling process on a large dataset, then divide it into smaller subsets that the GPU can handle.
We then build up the large lot using gradient accumulation.
It is crucial that we refrain from implementing any additional normalization operations that might change the gradient sensitivity \cite{ponomareva2023dp, Hoory.et.al.2021.FindingsEMNLP}, prior to the noise addition step.

\section{Experiments}

To demonstrate our framework in use, fill the gaps on current knowledge of the privacy/utility trade-off for the task of NMT, as well as examine the effects of using random shuffling vs.\ Poisson sampling, we run a series of experiments with DP-SGD on several NMT datasets, using a variety of privacy budgets.

\subsection{Datasets}

We utilize datasets comprising two main types of settings.
The first is the general NMT setting for comparing our models with previous work and investigating the effectiveness of DP-SGD on a common NMT dataset.
For this we utilize WMT-16 \citep{bojar2016findings}, using the German-English (DE-EN) language pair as the focus of our experiments.

The second setting is the more specific target domain of private texts that we are aiming to protect with differentially private NMT.
For the sake of reproducibility and ethical considerations, we utilize datasets that \textit{imitate} the actual private setting of processing sensitive information, namely business communications and medical notes, but are themselves publicly available. 
The first dataset is the Business Scene Dialogue corpus (BSD) \citep{rikters2019designing}, which is a collection of fictional business conversations in various scenarios (e.g.\ ``face-to-face'', ``phone call'', ``meeting''), with parallel data for Japanese and English.
While the original corpus consists of half English $\rightarrow$ Japanese and half Japanese $\rightarrow$ English scenarios, we combine both into a single Japanese $\rightarrow$ English (JA-EN) language pair for our experiments.

The second dataset is ClinSPEn-CC \citep{neves2022findings}, which is a collection of parallel COVID-19 clinical cases in English and Spanish, originally part of the biomedical translation task of WMT-22. We utilize this corpus in the Spanish $\rightarrow$ English (ES-EN) direction. 
These latter two datasets simulate a realistic scenario where a company or public authority may train an NMT model on private data, for later public use.
We present overall statistics for each dataset in Table~\ref{tab:dataset-statistics}.

\begin{table}[h]
    \centering
    \resizebox{0.5\textwidth}{!}{
    \begin{tabular}{lr|rr}
        \textbf{Dataset} & \textbf{Lang. Pair} & \textbf{\texttt{\#} Trn.+Vld.} & \textbf{\texttt{\#} Test} \\
        \hline
        WMT-16 & DE-EN & 4,551,054 & 2,999 \\
        BSD & JA-EN & 22,051 & 2,120 \\
        ClinSPEn-CC & ES-EN & 1,065 & 2,870 \\  
    \end{tabular}
    }
    \caption{Dataset statistics. Trn.: Train, Vld.: Validation.}
    \label{tab:dataset-statistics}
\end{table}

\subsection{Experimental setup}
\label{sec:exp-setup}

For each of the above three datasets, we fine-tune a pre-trained mT5 model \citep{xue-etal-2021-mt5}, opting for the \texttt{mT5-small}\footnote{\url{https://huggingface.co/google/mt5-small}} version due to computational capacity limitations described in Section~\ref{sec:methods}. 
We compare $\varepsilon$ values of $\infty, 1000, 5,$ and $1$, representing the non-private, weakly private, moderately private, and very private scenarios, respectively (see \citet{Lee.Clifton.2011.ISC,hsu2014differential,weiss2023share} for a more detailed discussion on selecting the `right' $\varepsilon$ value).
We fix the value of $\delta$ to $10^{-8}$ for all experiments, staying well below the recommended $\delta \ll \frac{1}{N}$ condition \citep{abadi2016deep}.

For all of the above configurations, we compare two methods of selecting batches of data points from the dataset for our DP-SGD configurations, namely \textbf{random shuffling} and \textbf{Poisson sampling}.
Following previous work \citep{Hoory.et.al.2021.FindingsEMNLP,anil-etal-2022-large,yin-habernal-2022-privacy}, we utilize very large batch sizes for both of these methods, setting $L$ to a large value and building up the resulting drawn batches with gradient accumulation for the latter method, as described in Section~\ref{sec:methods}.
We refer to Appendix~\ref{sec:appx-hyperparams} for a more detailed description of our hyperparameter search.
We evaluate our model outputs using BLEU \citep{papineni2002bleu}, and BERTScore \citep{zhang2019bertscore} metrics.

\subsection{Results and Discussion}
\label{sec:results}
\begin{figure}[!ht]
    \centering
    \includegraphics[trim={0 0 3cm 3.5cm},clip,width=\linewidth]{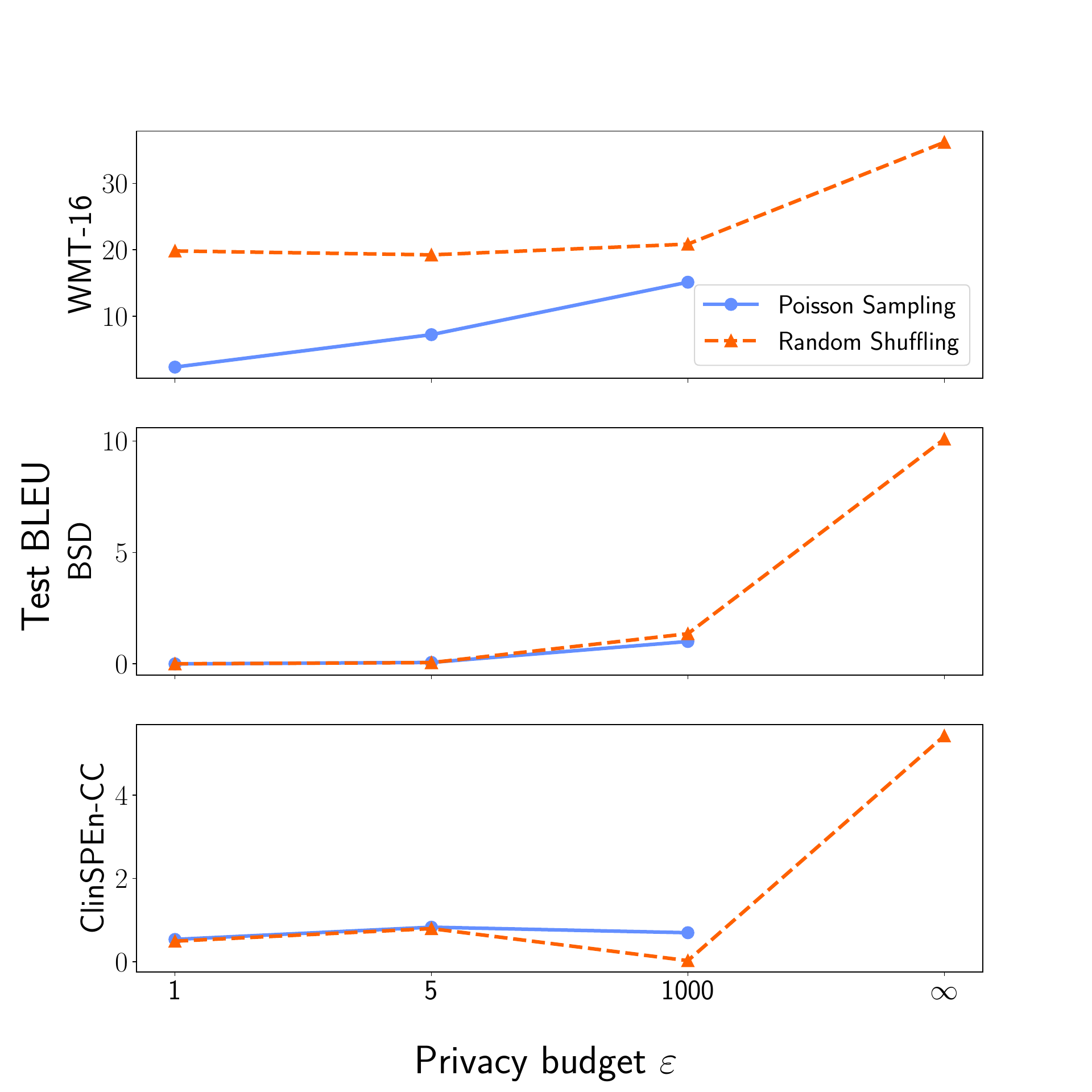}
    \caption{Test BLEU scores for each of the three datasets using varying privacy budgets, comparing the random shuffling and Poisson sampling methods to iterate over the dataset. Non-private results are additionally shown for each dataset ($\varepsilon = \infty$) with random shuffling. Lower $\varepsilon$ corresponds to a stronger privacy guarantee.}
    \label{fig:results}
\end{figure}

Figure~\ref{fig:results} shows the results of our experiments, reporting BLEU scores on the test partition of each dataset.

\paragraph{Privacy/utility trade-off}

We verify the soundness of our models in the non-private setting ($\varepsilon = \infty$) by comparing with past non-private results, particularly for the commonly used WMT-16 dataset.
For WMT-16 DE-EN, we reach a BLEU score of 36.2, being similar to past models (e.g.\ \citet{wei2021finetuned} obtain a BLEU score of 38.6 using their 137B parameter FLAN model).
In the case of BSD and ClinSPEn-CC, these datasets are not as `standard' within the NMT community, and therefore have a more limited chance for comparison.

For private results, we can see a clear difference between the drop in WMT-16 performance vs.\ that of BSD and ClinSPEn-CC.
This is not at all surprising, given that the latter two datasets are vastly smaller in comparison to WMT-16, making it far more difficult to train an NMT model, particularly in the noisy setting of DP-SGD.
In addition, ClinSPEn-CC contains a large amount of complicated medical terminology that adds an extra layer of difficulty for a model.
We therefore need to conduct further investigations into applications of DP-SGD to very small datasets in order to reach more meaningful $\varepsilon$ values. 

\paragraph{Method of dataset iteration}

When comparing random shuffling with Poisson sampling, we can see practically no difference for BSD and ClinSPEn-CC, most likely due to the low DP-SGD results for these two datasets.
The differences are more notable for WMT-16, where there is a clear gap between the two sets of configurations.
For instance, at $\varepsilon = 1$, WMT-16 shows a BLEU score of 19.83 when using random shuffling, in contrast to 2.35 with Poisson sampling.
The latter method therefore shows a far greater drop from the non-private setting, improving more gradually as $\varepsilon$ is increased.

There are several possible explanations for this.
With Poisson sampling, while each data point has an equal probability of being drawn to make up a particular batch, it is possible that some data points end up being drawn more frequently than others for several training iterations.
This may have an impact on the model learning process, possibly missing out on the signal from certain useful data points at various stages of training.
Another reason may be that we simply require additional hyperparameter optimization with Poisson sampling, expanding the search space further.

\section{Conclusion}

We have introduced \name, a modular framework developed using the JAX library, with the goal of leading research on neural machine translation with DP-SGD.
To demonstrate our framework in use, we have presented several experiments on both general and privacy-related NMT datasets, comparing two separate approaches for iterating over training data with DP-SGD, and facilitating in filling the research gap on the privacy/utility trade-off in this task. 
We are continuing to actively expand the framework, including the integration of new models and NMT datasets.
We hope that our framework will help to expand research into privacy-preserving NMT and welcome feedback from the community.

\section*{Ethics and Limitations}
\label{sec:ethics-and-limitations}

An important ethical consideration with regards to our framework is its intended use.
We strive to further the field of private NMT and improve the current knowledge on how to effectively apply differential privacy to data used in NMT systems.
However, applications of differential privacy to textual data are still at an early research stage, and \textbf{should not currently be used in actual services that handle real sensitive data of individuals.}

The primary reason for this is that our understanding of what is \textit{private information} in textual data is still very limited.
Applications of differential privacy in the machine learning setting provide a privacy guarantee to each individual \textit{data point}.
In the context of DP-SGD, this means that if any single data point is removed from the dataset, the impact on the resulting model parameter update is bounded by the provided multiplicative guarantee in Eqn.~\ref{dp-definition}.
In other words, it does not stand out `too much' in its contribution to training the model.

For textual data, a single data point will often be a sentence or document.
However, this does not mean that there is a one-to-one mapping from \textit{individuals} to sentences and documents.
For instance, multiple documents could potentially refer to the same individual, or contain the same piece of sensitive information that would break the assumption of each data point being independent and identically distributed (i.i.d.) in the DP setting.
Thus, we require further research on how to properly apply a privacy guarantee to individuals represented within a textual dataset. 
We refer to \citet{klymenko-etal-2022-differential,brown2022does,igamberdiev-habernal-2023-dp} for a more comprehensive discussion on this.

\section*{Acknowledgements}

This project was supported by the PrivaLingo research grant (Hessisches Ministerium des Innern und für Sport).
The independent research group TrustHLT is supported by the Hessian Ministry of Higher Education, Research, Science and the Arts. 
Thanks to Luke Bates for helpful feedback on a preliminary draft.

\bibliography{custom}
\bibliographystyle{acl_natbib}

\appendix

\section{Background on Differential Privacy and DP-SGD}
\label{sec:appx-dp-background}

\paragraph{Differential Privacy}
Differential privacy (DP) is a mathematical framework which formally guarantees that the output of a randomized algorithm $\mathcal{M} : \mathcal{X} \rightarrow \mathcal{Y}$ abides by the following inequality in Eqn.~\ref{dp-definition}, for all \textit{neighboring} datasets $x, x' \in \mathcal{X}$, i.e.\ datasets which are identical to one another, with the exception of one data point \citep{Dwork.Roth.2013}
\begin{equation}
\Pr[\mathcal{M}(x) \in S] \leq e^{\varepsilon}\Pr[\mathcal{M}(x') \in S] + \delta,
\label{dp-definition}
\end{equation}
for all $S \subseteq \mathcal{Y}$.

We refer to the algorithm $\mathcal{M}$ as being ($\varepsilon, \delta$)-differentially private, where $\varepsilon \in [0, \infty)$, also known as the \textit{privacy budget}, represents the strength of the privacy guarantee. A lower $\varepsilon$ value represents an exponentially stronger privacy protection. $\delta \in [0, 1]$ is a very small constant which relaxes the pure differential privacy of ($\varepsilon, 0$)-DP, providing better composition when iteratively applying multiple DP mechanisms to a given dataset.

In order to transform a non-private algorithm $f : \mathcal{X} \rightarrow \mathcal{Y}$ into one satisfying an ($\varepsilon, \delta$)-DP guarantee, we generally add Gaussian noise to the output of $f$. Overall, the whole process restricts the degree to which any single data point can stand out when applying algorithm $\mathcal{M}$ on a dataset.

\paragraph{DP-SGD}
A popular method for applying DP to the domain of machine learning is through differentially private stochastic gradient descent (DP-SGD) \citep{abadi2016deep}.
The core of the methodology relies on adding two extra steps to the original stochastic gradient descent algorithm. For any input data point $x_i$, we first calculate the gradient of the loss function for a model with parameters $\theta$, $\mathcal{L}(\theta)$, at training iteration $t$. Hence, $g_t(x_i) = \nabla_{\theta_t} \mathcal{L}(\theta_t, x_i)$.

We then incorporate a \textit{clipping} step, in which the $\ell_2$-norm of $g_t(x_i)$ is clipped with clipping constant $C$, as in Eqn.~\ref{dp-sgd-clipping}, in order to constrain the range of possible values.
This is followed by a \textit{perturbation} step, adding Gaussian noise to the clipped gradients, as in Eqn.~\ref{dp-sgd-perturbation}.

\begin{equation}
\bar{g}_t(x_i) = \frac{g_t(x_i)}{\max \left( 1, \frac{||g_t(x_i)||_2}{C} \right)}
\label{dp-sgd-clipping}
\end{equation}

\begin{equation}
\hat{g}_t = \frac{1}{L} \sum_{i \in L} \left(  \bar{g}_t(x_i) + \mathcal{N}(0, \sigma^2C^2\mathbf{I}) \right)
\label{dp-sgd-perturbation}
\end{equation}
Importantly, $L$ represents the \textit{lot size}, being a group of data points that are randomly drawn from the full training dataset at each iteration.
The final gradient descent step is then taken with respect to this noisy gradient $\hat{g}_t$. 
We outline the DP-SGD algorithm in more detail in Algorithm~\ref{alg:dp-sgd}.

\begin{algorithm}[h!]
    \caption{DP-SGD}
    \label{alg:dp-sgd}
    \begin{algorithmic}[1]
        \Function{DP-SGD}{$f(\bm{x}; \Theta)$, $(\bm{x}_1, \ldots, \bm{x}_n)$, $|L|$ --- `lot' size, $T$ --- \# of steps}
        \For{$t \in (1, 2, \ldots, T)$}
        \State Add each training example to a `lot' $L_t$ with probability $|L|/N$
        \For{each example in the `lot' $\bm{x}_i \in L_t$}
        \State $\bm{g}(\bm{x}_i) \gets \nabla \mathcal{L} (\theta_t, \bm{x}_i)$
        \Comment{Compute gradient}
        \State
        $\bar{\bm{g}}(\bm{x}_i) \gets \bm{g}(\bm{x}_i) / \max \left(1 , \|  \bm{g}(\bm{x}_i) \| / C \right)$
        \Comment{Clip gradient}
        \State $\tilde{\bm{g}}(\bm{x}_i) \gets \bar{\bm{g}}(\bm{x}_i) + \mathcal{N}(0, \sigma^2 C^2 \bm{I})$
        \Comment{Add noise}
        \EndFor
        \State $\hat{\bm{g}} \gets \frac{1}{|L|} \sum_{k = 1}^{|L|} \tilde{\bm{g}}(\bm{x}_k)$
        \Comment{Gradient estimate of `lot' by averaging}
        \State $\Theta_{t + 1} \gets \Theta_t - \eta_t \hat{\bm{g}}$
        \Comment{Update parameters by gradient descent}
        \EndFor
        \State \Return $\Theta$
        \EndFunction
    \end{algorithmic}
\end{algorithm}

\section{Hyperparameters}
\label{sec:appx-hyperparams}

We present our hyperparameter search space as follows.
We experiment with learning rates in the range $[10^{-5}, 0.01]$ and maximum sequence lengths in $[8, 64]$.
Following previous work, we utilize large batch and lot sizes for our experiments, finding $1,048,576$ to be the best for WMT-16, $2,048$ for BSD, and $256$ for ClinSPEn-CC.
We build up these batch sizes using gradient accumulation with a physical batch size of $16$.
In the case of Poisson sampling, we first sample using large lot sizes and build the resulting drawn batch using gradient accumulation, as described in Section~\ref{sec:methods}.
We train models for up to $25$ epochs, using the same definition for \textit{epochs} as in \citet{abadi2016deep} in the Poisson sampling setting, being $\frac{N}{L}$. We take the ceiling in case of $L$ not cleanly dividing into $N$.
Each configuration is run using 5 seeds for the BSD and ClinSPEn-CC datasets and 3 seeds for WMT-16, reporting the mean and standard deviation of results.

We additionally present our computational runtimes in Table~\ref{tab:computational-runtimes}.
All experiments are run on up to two 80GB NVIDIA A100 Tensor Core GPUs.

\begin{table}[h]
    \centering
    \resizebox{0.5\textwidth}{!}{
    \begin{tabular}{lrr|r}
        \textbf{Dataset} & \textbf{$\varepsilon$} & \textbf{Iteration Method} & \textbf{Epoch Time} \\
        \hline
        WMT-16 & $\infty$ & Random shuffling & 2 h 45 m 08 s \\
        WMT-16 & $1000$ & Random shuffling & 2 h 59 m 15 s \\
        WMT-16 & $1000$ & Poisson sampling & 4 h 08 m 01 s \\
        WMT-16 & $5$ & Random shuffling & 1 h 30 m 03 s \\
        WMT-16 & $5$ & Poisson sampling & 4 h 02 m 35 s \\
        WMT-16 & $1$ & Random shuffling & 1 h 29 m 49 s \\
        WMT-16 & $1$ & Poisson sampling & 4 h 09 m 02 s \\
        BSD & $\infty$ & Random shuffling & 0 h 01 m 17 s \\
        BSD & $1000$ & Random shuffling & 0 h 01 m 59 s \\
        BSD & $1000$ & Poisson sampling & 0 h 01 m 49 s \\
        BSD & $5$ & Random shuffling & 0 h 00 m 52 s \\
        BSD & $5$ & Poisson sampling & 0 h 01 m 49 s \\
        BSD & $1$ & Random shuffling & 0 h 01 m 09 s \\
        BSD & $1$ & Poisson sampling & 0 h 02 m 15 s \\
        ClinSPEn-CC & $\infty$ & Random shuffling & 0 h 00 m 09 s \\
        ClinSPEn-CC & $1000$ & Random shuffling & 0 h 00 m 05 s \\
        ClinSPEn-CC & $1000$ & Poisson sampling & 0 h 00 m 28 s \\
        ClinSPEn-CC & $5$ & Random shuffling & 0 h 00 m 10 s \\
        ClinSPEn-CC & $5$ & Poisson sampling & 0 h 00 m 27 s \\
        ClinSPEn-CC & $1$ & Random shuffling & 0 h 00 m 15 s \\
        ClinSPEn-CC & $1$ & Poisson sampling & 0 h 00 m 27 s \\
    \end{tabular}
    }
    \caption{Sample epoch runtimes for each configuration. Some differences between configurations arise due to different optimal hyperparameters, with larger sequence lengths leading to longer epoch times.} 
    \label{tab:computational-runtimes}
\end{table}

\section{Detailed Results}
\label{sec:appx-detailed-results}

\begin{table*}[h]
    \centering
    \resizebox{0.7\textwidth}{!}{
    \begin{tabular}{lrr|rr}
        \textbf{Dataset} & \textbf{$\varepsilon$} & \textbf{Iteration Method} & \textbf{Test BLEU} & \textbf{Test BERTScore} \\
        \hline
        WMT-16 & $\infty$ & Random shuffling & 36.19 (0.13) & 0.95 (0.00) \\
        WMT-16 & $1000$ & Random shuffling & 20.86 (0.56) & 0.92 (0.00) \\
        WMT-16 & $1000$ & Poisson sampling & 15.12 (0.08) & 0.91 (0.00) \\
        WMT-16 & $5$ & Random shuffling & 19.24 (0.52) & 0.92 (0.00) \\
        WMT-16 & $5$ & Poisson sampling & 7.23 (0.21) & 0.89 (0.00) \\
        WMT-16 & $1$ & Random shuffling & 19.83 (0.64) & 0.92 (0.00) \\
        WMT-16 & $1$ & Poisson sampling & 2.35 (0.07) & 0.84 (0.00) \\
        BSD & $\infty$ & Random shuffling & 10.09 (2.75) & 0.90 (0.01) \\
        BSD & $1000$ & Random shuffling & 1.36 (0.67) & 0.87 (0.01) \\
        BSD & $1000$ & Poisson sampling & 1.01 (0.07) & 0.87 (0.00) \\
        BSD & $5$ & Random shuffling & 0.06 (0.05) & 0.85 (0.01) \\
        BSD & $5$ & Poisson sampling & 0.06 (0.06) & 0.84 (0.02) \\
        BSD & $1$ & Random shuffling & 0.00 (0.01) & 0.45 (0.22) \\
        BSD & $1$ & Poisson sampling & 0.00 (0.00) & 0.65 (0.15) \\
        ClinSPEn-CC & $\infty$ & Random shuffling & 5.42 (2.41) & 0.86 (0.02) \\
        ClinSPEn-CC & $1000$ & Random shuffling & 0.03 (0.02) & 0.75 (0.01) \\
        ClinSPEn-CC & $1000$ & Poisson sampling & 0.70 (0.19) & 0.78 (0.00) \\
        ClinSPEn-CC & $5$ & Random shuffling & 0.80 (0.56) & 0.79 (0.00) \\
        ClinSPEn-CC & $5$ & Poisson sampling & 0.83 (0.27) & 0.79 (0.00) \\
        ClinSPEn-CC & $1$ & Random shuffling & 0.50 (0.20) & 0.78 (0.00) \\
        ClinSPEn-CC & $1$ & Poisson sampling & 0.54 (0.22) & 0.78 (0.00) \\
    \end{tabular}
    }
    \caption{Detailed results of each experimental configuration. Scores shown as ``mean (standard deviation)''. Results show the average over 3 seeds for the WMT-16 dataset, and 5 seeds for BSD and ClinSPEn-CC.}
    \label{tab:detailed-results}
\end{table*}

\end{document}